\documentclass[sigconf]{acmart}

\AtBeginDocument{%
  }

\usepackage{xspace}
\usepackage{float}
\usepackage{algorithm}
\usepackage{algpseudocode}
\usepackage{todonotes}
\usepackage{booktabs}

\algrenewcommand\algorithmicrequire{\textbf{Input:}}
\algrenewcommand\algorithmicensure{\textbf{Output:}}

\usepackage{listings}
\usepackage{xcolor}

\definecolor{codegray}{rgb}{0.5,0.5,0.5}
\definecolor{codepurple}{rgb}{0.58,0,0.82}
\definecolor{codegreen}{rgb}{0,0.6,0}
\definecolor{backcolour}{rgb}{0.95,0.95,0.95}

\lstdefinelanguage{json}{
  morestring=[b]",
  morecomment=[l]{//},
  stringstyle=\color{codegreen},
  commentstyle=\color{codegray},
  literate=
   *{0}{{{\color{codepurple}0}}}{1}
    {1}{{{\color{codepurple}1}}}{1}
    {2}{{{\color{codepurple}2}}}{1}
    {3}{{{\color{codepurple}3}}}{1}
    {4}{{{\color{codepurple}4}}}{1}
    {5}{{{\color{codepurple}5}}}{1}
    {6}{{{\color{codepurple}6}}}{1}
    {7}{{{\color{codepurple}7}}}{1}
    {8}{{{\color{codepurple}8}}}{1}
    {9}{{{\color{codepurple}9}}}{1}
}

\lstdefinestyle{jsonstyle}{
  language=json,
  basicstyle=\ttfamily\footnotesize,
  keywordstyle=\color{codepurple},
  numbers=left,
  numberstyle=\tiny\color{white},
  stepnumber=1,
  numbersep=5pt,
  showstringspaces=false,
  breaklines=true,
  frame=none,
  tabsize=2,
  captionpos=b
}

\copyrightyear{2026}
\acmYear{2026}
\setcopyright{cc}
\setcctype{by}
\acmConference[WWW '26]{Proceedings of the ACM Web Conference 2026}{April 13--17, 2026}{Dubai, United Arab Emirates}
\acmBooktitle{Proceedings of the ACM Web Conference 2026 (WWW '26), April 13--17, 2026, Dubai, United Arab Emirates}
\acmDOI{10.1145/3774904.3792988}
\acmISBN{979-8-4007-2307-0/2026/04}

\newcommand{\tbar}{\texttt{\text{T-BARS}}\xspace}
\newcommand{\co}{\texttt{\text{coTherapist}}\xspace}

\begin{document}

\title{coTherapist: A Behavior-Aligned Small Language Model to Support Mental Healthcare Experts}

\author{Prottay Kumar Adhikary}
\affiliation{%
  \institution{IIT Delhi}
  \city{New Delhi}
  \country{India}
  }
\email{prottay71@gmail.com}

\author{Reena Rawat}
\affiliation{%
  \institution{IIT Delhi}
  \city{New Delhi}
  \country{India}
  }
\email{reenarawat755@gmail.com}

\author{Tanmoy Chakraborty}
\affiliation{
\institution{IIT Delhi, India}
\institution{IIT Delhi Abu Dhabi, UAE}
\country{}
}
\email{tanchak@iitd.ac.in}

\begin{abstract}
Access to mental healthcare is increasingly strained by workforce shortages and rising demand, motivating the development of intelligent systems that can support mental healthcare experts. We introduce \co, a unified framework utilizing a small language model to emulate core therapeutic competencies through domain-specific fine-tuning, retrieval augmentation, and agentic reasoning. Evaluation on clinical queries demonstrates that \co\ generates more relevant and clinically grounded responses than contemporary baselines. Using our novel \tbar\ rubric and psychometric profiling, we confirm \co exhibits high empathy and therapist-consistent personality traits. Furthermore, human evaluation by domain experts validates that \co\ delivers accurate, trustworthy, and safe responses. \co was deployed and tested by clinical experts. Collectively, these findings demonstrate that small models can be engineered to exhibit expert-like behavior, offering a scalable pathway for digital mental health tools.
\end{abstract}



\maketitle

\begin{figure}[t]
  \centering
  \includegraphics[width=\columnwidth]{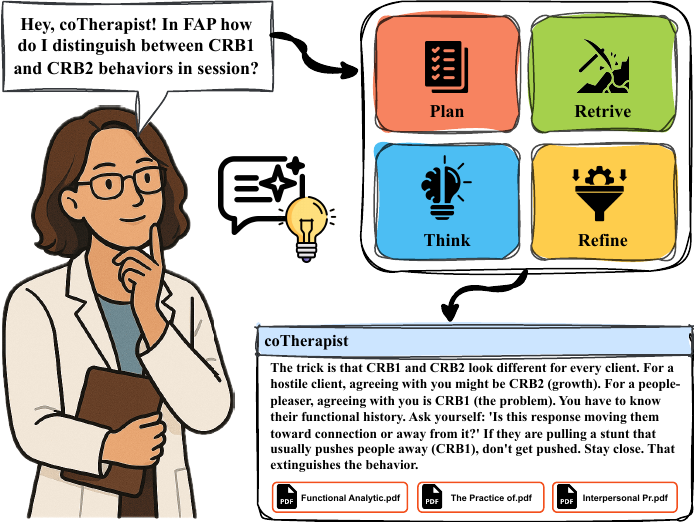}
    \caption{Illustrative interaction with \co. The figure visualizes the end-to-end pipeline triggered by a mental healthcare expert's query, highlighting the stages of \textit{planning}, \textit{retrieval}, \textit{internal reasoning}, and \textit{self-refinement}. The final output demonstrates how the system generates a response that is grounded in cited sources and aligned with a professional therapeutic tone, despite being powered by a lightweight small language model.}
  \label{fig:cotherapist-flow}
  \vspace{-2mm}
\end{figure}

\section{Introduction}
Digital mental health platforms are increasingly utilizing Artificial Intelligence (AI) to bridge gaps in care and support Mental Health Experts (MHx). Large Language Models (LLMs) have recently demonstrated a surprising ability to provide critical advice \cite{huang2025aiaugmentedllmsachievetherapistlevel}. While these systems promise better accessibility and can offer support that users find clearer and less biased than human therapists \cite{James2024, Yang2024}, naive implementation can still lead to unhelpful, script-like errors or harmful outputs \cite{iftikhar2025therapynlptaskpsychologists}. To address this within the emerging field of computational psychiatry \cite{Dehghan2025}, it is essential to define an AI's ``behavior'' not merely as text generation, but as its observable communication patterns \cite{Louwen2023}. For a model to truly resemble a skilled therapist rather than a generic chatbot, it must master three specific components: style imitation involving tone and paraphrasing \cite{Soma2021}, a structured instruction-following capability \cite{Webb2010}, and a conceptual reasoning framework based on evidence-based therapy logic \cite{Wilcox2023}.

\paragraph{AI as a Therapeutic Assistant}
Rather than attempting to replace human clinicians, we envision AI serving as a supportive assistant that helps trained practitioners deliver better care \cite{10.1145/3613904.3642443}. In clinical practice, therapists often consult manuals, worksheets, or peers to guide their interventions \cite{Tong2025}, but they generally avoid relying on the open internet due to the high risk of misinformation. Similarly, an AI could fulfill this role by instantly answering a therapist’s query about treatment techniques, including specific book sections for verification. It can also provide the rationale for a diagnosis or suggest a gentle way to rephrase a question asked to a client \cite{zhang_can_2024}. By doing so, it can ease the therapist’s cognitive load and provide quick access to consolidated expertise, all while the therapist remains in charge of the client’s care \cite{Lee2021}. This human-AI collaboration approach, which we term \textit{coTherapy}, aligns with the broader goals of digital well-being, leveraging web-based platforms to improve mental health support without removing the human element \cite{Olawade2024}.

\paragraph{Challenges with Small Language Models}
Transforming a general Small Language Model (SLM) into an effective clinical tool requires the model to adopt core counseling skills like reflective listening and structured frameworks such as Cognitive Behavioral Therapy (CBT) \cite{chiu2024bolt}. While generic SLMs are proficient at following technical rules, they often fall short in relational subtleties, producing formulaic empathy rather than the nuanced validation provided by human experts \cite{arxiv.2403.06609}. To bridge this gap, we need both \textit{behavioral style alignment} and \textit{conceptual alignment}, enabling the AI to ``think through'' problems using psychological theories.

\paragraph{Our Contributions.} We introduce \co, a lightweight agent emulating expert therapeutic behavior \cite{Kim2025} (Figure \ref{fig:cotherapist-flow}). Designed strictly as a clinician's assistant, not a replacement, it is evaluated on simulated queries. Unlike computationally intensive LLMs \cite{Singhal2023}, we focus on aligning SLMs with clinical standards \cite{arxiv.2507.08031, arxiv.2306.11644}, enabling deployment in resource-constrained settings \cite{Wubineh2024, Xu2020}. Prioritizing behavioral fidelity over raw performance, our contributions are: {\bf (1) Domain-Specific Psychotherapy Knowledge Dataset.} We curated a dataset of 800 million+ tokens, comprising undergraduate and postgraduate textbooks, lecture notes and videos, practice materials, etc. This corpus grounds the model in evidence-based therapeutic language, conceptual reasoning, and technical accuracy. {\bf (2) The \co Framework.} We develop a multi-stage pipeline for a 1 Billion parameter SLM, integrating continued pretraining for domain adaptation, LoRA fine-tuning for dialogue alignment \cite{Faridi2025}, and Retrieval-Augmented Generation (RAG) for evidentiary support \cite{Shah2025}. Additionally, we implement an agentic reasoning structure to facilitate step-by-step clinical formulation. {\bf (3) Therapist Behavior Rating Scale.} We propose \tbar, a novel framework assessing four key pillars: Behavioral Style, Conceptual Reasoning, Relational Competence, and Technique Execution, across 20 sub-skills. Using LLM-based judges and psychometric profiling \cite{li-etal-2025-big5}, we find that \co achieves a 3 improvement in empathy and relational clarity over baselines, despite modest gains in standard metrics. Human evaluation further confirms expert preference for \co's safety and helpfulness.\footnote{We are committed to releasing the source codes and dataset upon acceptance of the paper.}

\section{Related Work}
The application of NLP in mental health has evolved from simple rule-based chatbots to complex generative systems \cite{Vaidyam2019}. Recent advancements demonstrate that LLMs can be aligned with professional therapeutic standards \cite{10.1145/3564752}. For instance, models fine-tuned for CBT have shown significantly higher adherence to counseling strategies compared to base models \cite{Kim2025}. Beyond behavioral adherence, research indicates that LLMs can infer latent psychological structures, accurately reasoning about personality patterns based on Big Five traits \cite{liu2025dimensionsmanylargelanguage}.

This capability is essential for developing a consistent ``therapist persona,'' shifting the field's focus from client-facing chatbots to AI assistants that support clinical decision-making, documentation, and retrieval of evidence-based criteria. Despite progress in generative capabilities, there remains limited research on designing AI systems specifically to reduce a practitioner's need to constantly consult external manuals or textbooks during practice \cite{Rebelo2023}. Generic models often struggle with this due to hallucinations. To mitigate this, RAG has emerged as a standard for grounding LLMs \cite{Gams2025} in external knowledge bases \cite{arxiv.2005.11401}, a technique we adapt specifically for psychotherapy manuals \cite{Son2025}. 

Turning a general SLM into an effective AI requires the model to adopt core counseling skills and clinical reasoning patterns. Generic SLMs often fall short: prior research comparing AI counselors to humans found that while SLMs are good at adhering to formal techniques, they struggle with relational subtleties, sometimes giving formulaic empathy or over-direct advice \cite{arxiv.2403.06609}. For example, an LLM might correctly suggest an exposure therapy exercise but fail to respond to a client's emotional cues with appropriate validation.

In clinical settings, Chain-of-Thought (CoT) mirrors the diagnostic reasoning process of a human therapist \cite{arxiv.2408.06527}. While some studies have applied CoT to improve diagnostic accuracy, few have applied it to interpersonal alignment, specifically \cite{wang2023cuecot}, reasoning about how to phrase a response therapeutically (e.g., ``\textit{The client seems anxious; I should use validation before suggesting a technique}''). By integrating an internal reasoning layer, models can move beyond surface-level style imitation to emulate the cognitive process of a junior therapist \cite{Kim2025, liu2025dimensionsmanylargelanguage}.

\begin{table*}[!t]
  \centering
  \caption{Schematic overview of therapeutic modality clusters and their corresponding disorder coverage. This mapping illustrates the breadth of clinical interventions integrated into the knowledge base, categorizing evidence-based frameworks (e.g., CBT, DBT, ACT) alongside the specific psychopathological conditions and symptom profiles they are primarily designed to treat.}
  \label{tab:modality_coverage}
  \begin{tabular}{p{8.5cm}p{8.5cm}}
    \toprule
    \textbf{Modality Cluster} \& \textbf{Primary Disorder Coverage} \\
    \midrule

    Cognitive Behavioral Therapy family (Cognitive Behavioral Therapy (CBT), Cognitive Therapy (CT), Rational Emotive Behavior Therapy (REBT), Rumination-Focused Cognitive Behavioral Therapy (RFCBT), Mindfulness-Integrated Cognitive Behavioral Therapy (MiCBT), Exposure and Response Prevention (EX/RP or ERP)) &
    Depression, anxiety spectrum (Generalized Anxiety Disorder (GAD), panic disorder), Obsessive–Compulsive Disorder (OCD), Post-Traumatic Stress Disorder (PTSD) / trauma, Body Dysmorphic Disorder (BDD), Family/domestic violence, Anger, Rumination. \\
    \midrule

    Dialectical Behavior Therapy family (Dialectical Behavior Therapy (DBT), skills-based mindfulness, emotion regulation methods) &
    Borderline Personality Disorder (BPD) / emotion dysregulation, Suicidality/crisis, Comorbid Personality Disorders (PD) + Substance Use Disorders (SUD) / eating disorders, Overwhelming emotions. \\
    \midrule

    Acceptance and Commitment Therapy / Mindfulness family (Acceptance and Commitment Therapy (ACT), Mindfulness-Based Cognitive Therapy (MBCT), Mindfulness-Based Cognitive Behavioral Therapy (MBCBT)) &
    Stress, anxiety, depression, Relapse prevention, Experiential avoidance, Mixed transdiagnostic problems. \\
    \midrule

    Interpersonal / Rhythm / Family systems (Interpersonal Therapy (IPT), Interpersonal and Social Rhythm Therapy (IPSRT), Family/Couple therapies) &
    Depression, Bipolar Disorder (BD), Eating disorders, PTSD, Marital/relational distress, Family violence. \\
    \midrule

    Trauma-focused (Eye Movement Desensitization and Reprocessing (EMDR), Somatic therapy, Emotion-Focused Therapy (EFT) trauma applications) &
    PTSD, Sexual trauma, Complex trauma, Trauma-related anxiety and depression. \\
    \midrule

    Behavioural / Developmental (Applied Behavior Analysis (ABA), Parent Management Training (PMT), Treatment and Education of Autistic and Communication-Handicapped Children (TEACCH), Behavior modification) &
    Autism Spectrum Disorder (ASD), Attention-Deficit/Hyperactivity Disorder (ADHD), Tourette Syndrome (TS), Oppositional Defiant Disorder (ODD) / conduct problems / aggression, Intellectual Disability (ID) / Developmental Disability (DD), Child/adolescent behavior disorders. \\
    \midrule

    Integrative / Supportive / Existential / Logotherapy / Psychodrama &
    Broad distress, Neuroses, Psychosomatic/functional illness, Severe mental illness support, Life challenges. \\
    \midrule

    Sex therapy (Biopsychosocial / Systemic / Integrative approaches) &
    Premature Ejaculation (PE), Erectile Disorder (ED), Orgasmic disorders, Desire/arousal issues, Genito-pelvic pain disorders, Paraphilic disorders, Gender Dysphoria (GD). \\
    \bottomrule
  \end{tabular}
\end{table*}

\section{Dataset: Psychotherapy Knowledge Corpus}
Our dataset is a curated psychotherapy knowledge corpus \textbf{(PsyKC)} designed to (i) provide authoritative clinical content for retrieval, and (ii) expose the model to the pedagogical and procedural language used in therapist training. The corpus pools publicly available multiple content families: therapy manuals, clinical psychology texts, lecture video transcripts, psychiatry references, diagnostic guidelines, and supporting disciplines (abnormal psychology, neuro-psychology, psychosocial interventions, social psychology, and research methods). Each item in the catalog is annotated with three special metadata fields: \textit{primary\_topic} (source family), \textit{therapeutic\_modality} (e.g., CBT, Dialectical Behavior Therapy (DBT), Eye Movement Desensitization and Reprocessing (EMDR), and \textit{specific\_disorder} (target clinical condition). Table \ref{tab:modality_coverage} shows the breadth of modality and disorder pairings. We merge duplicates during indexing (e.g., repeated CBT textbooks) so that each source contributes once to the vector store. To facilitate consistent evaluation, an expert curated a benchmark of 100 question-answer pairs, which was used to test the model across all experimental stages.

\textbf{\textit{Scale and composition.}}
In total, the knowledge base contains:
(i) \textbf{311 therapy and psychiatry books/manuals} ($\sim$2 GB raw text, $\sim$524M tokens after extraction), spanning both evidence-based protocols and integrative/eclectic approaches.
    (ii)  \textbf{250 lecture notes and training handouts, 552 lecture video transcripts} from graduate counseling/clinical programs, capturing ``how clinicians teach clinicians'' language, contributing another $\sim$227 million tokens.
    (iii) \textbf{121 Diagnostic and practice guidelines} emphasizing standardized assessment, risk management, and care pathways, adding $\sim$49 million tokens more.
This multi-family structure ensures the model has access to (a) {procedural steps} (manuals, worksheets), (b) {conceptual framing} (clinical/abnormal psychology), (c) {diagnostic grounding} (psychiatry texts and guidelines), and (d) \emph{assessment literacy} (neuro-psychology and research methods). This covers high-priority conditions including depression, the anxiety spectrum, trauma, personality disorders, psychosis, substance use, and neuro-developmental issues. It also integrates essential cross-cutting areas like crisis prevention, gender-affirming care, domestic violence support, and case conceptualization.\\

\textbf{\textit{Why this structure matters?}}
Therapy manuals and books provide the highest density of actionable techniques (e.g., exposure hierarchies, grounding scripts), while clinical psychology and psychiatry texts provide diagnostic nuance and case-formulation reasoning \cite{10.1145/3613904.3642761}. Each tier serves a distinct purpose. Guidelines anchor safety-critical behaviors \cite{Saeidnia2024} such as suicide risk pathways and evidence-based service models \cite{Iftikhar2025}. This layered design supports our RAG pipeline: retrieval surfaces \emph{exact} protocol text, while fine-tuning aligns the model’s tone and instructional ability. By mirroring the curriculum of formal clinical training \cite{Ni2025}, this diverse corpus ensures the model moves beyond rote procedural recall to acquire the underlying ``clinical logic'' necessary for adapting interventions to specific client needs.

\begin{figure*}[!t]
    \centering
    \includegraphics[width=\textwidth]{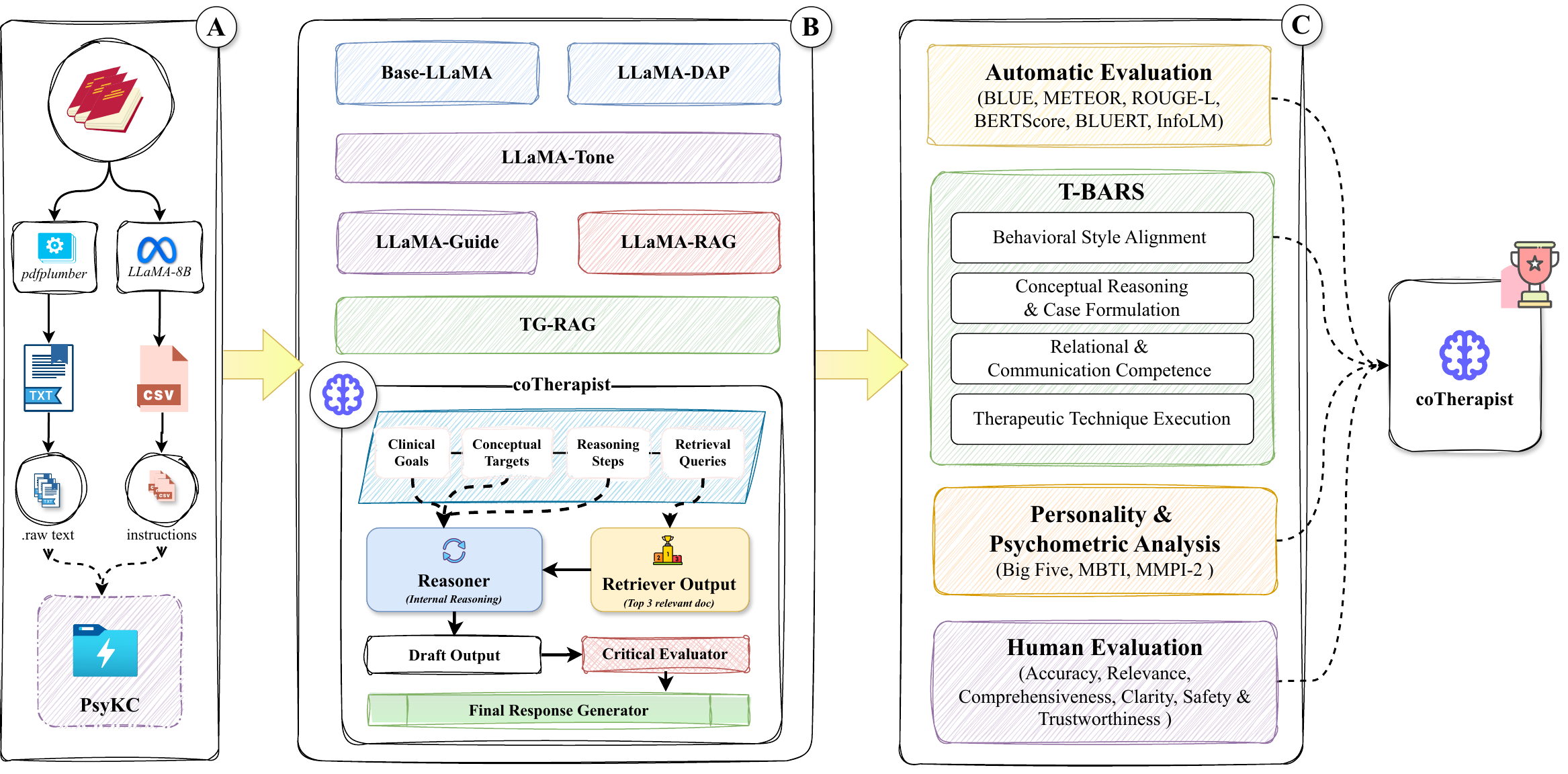}
   \caption{
    System overview of the proposed \co framework.  
    \textbf{(A) Data Curation:} We compile a high-quality clinical corpus from psychotherapy textbooks, university lecture notes, lecture videos, etc. The data is cleaned, segmented, and indexed to cover major evidence-based therapies used in MHx training and practice. 
    \textbf{(B) Experiment Design:} The model is trained through continued pretraining for domain tone, LoRA fine-tuning for therapist communication style, and retrieval-augmented generation to provide clinically accurate knowledge. An agentic reasoning step supports structured and context-aware responses.  
    \textbf{(C) Evaluation:} Outputs are evaluated using both traditional NLG metrics and the T-BARS behavioral framework. Across all settings, the full \co model demonstrates the highest therapist-like alignment and is preferred in human evaluation.
}
    \label{fig:cotherapist-full}
\end{figure*}

\section{Methodology}
\co\ integrates few components: model fine-tuning, retrieval architecture, and agentic reasoning, to build a psychotherapy-aware assistant that is both clinically-grounded and conversationally-aligned with therapist communication styles. We employ \textit{LLaMA 3.2-1B-Instruct}, the smallest model from the Meta family, as our base to facilitate accessible edge deployment, overcoming the inherent limitations of small models through a staged pipeline of domain-adaptive pretraining, stylistic LoRA fine-tuning, and self-instruction. This framework is further enhanced by a RAG layer and an internal reasoning and critique architecture, enabling the model to draw on authoritative protocols while maintaining human-like interaction. To ensure consistent adherence to evidence-based practices and prevent stylistic drift, all model variants utilize a unified base system prompt (Appendix \ref{app:system-prompt}) that instructs the models to act as a clinically grounded co-therapist. Our experimental design evaluates a few incremental approaches, starting from the base model and progressively adding these engineering techniques to culminate in the full \co\ framework. Figure \ref{fig:cotherapist-full} illustrates the complete \co pipeline from clinical data curation to model alignment and multi-stage evaluation.

\subsection*{Domain-Adaptive Pretraining (DAP)}
We perform unsupervised domain-adaptive pretraining \cite{gururangan-etal-2020-dont} on our collected corpus to infuse domain-specific terminology. Training on all textbook text for five epochs using the HuggingFace Trainer (batch size 128, sequence length 1024 on 1$\times$A100 GPU, learning rate $1\times10^{-5}$) \cite{wolf-etal-2020-transformers} yields {\bf LLaMA-DAP}. This model demonstrates improved clinical precision; for instance, discussing ``behavioral activation'' elicits terms like ``inhibitory learning'' rather than generic language (see Appendix \ref{appendix}). Early stopping is employed to prevent overfitting on domain text \cite{Xu2024}.

\subsection*{LoRA Fine-Tuning for Style}
To address DAP's didactic tone, we fine-tune LLaMA-DAP on therapist turns from the MentalCLOUDS benchmark \cite{Adhikary2024} (a state-of-the-art benchmark in mental health) using Low-Rank Adaptation (LoRA) ($r=16, \alpha=32, \text{dropout}=0.05$) \cite{arxiv.2106.09685}. This yields {\bf LLaMA-Tone}, which exhibits human-like therapeutic qualities such as first-person empathy \cite{arxiv.2309.15461}. We observe that this efficient style alignment does not significantly degrade the base knowledge.

\subsection*{Self-Instruction Tuning}
To enhance structured guidance, we employ the self-instruct paradigm \cite{selfinstruct}, generating $\sim$24k therapy-specific instruction pairs using \text{\textit{LLama-3.1-8B}}. Fine-tuning on this data produces \textbf{LLaMA-Guide}, which improves reasoning but diminishes conversational warmth \cite{arxiv.2304.12244}. An additional tuning round atop LLaMA-Tone balances these traits, yielding the final \textbf{LLaMA-TG} model.

\begin{table*}[!t]
\caption{Overview of incremental development stages and their resulting qualitative behaviors. The comparison illustrates how a small model shifts behavior from a generic to a clinically nuanced, empathetic co-therapist.}
  \label{tab:model_approaches}
  \begin{tabular}{p{3.2cm}p{3.3cm}p{10.2cm}}
    \toprule
    \textbf{Model Variant} & \textbf{Description} & \textbf{Observed Behavior}\\
    \midrule
    Base \\(\textit{LLaMA-3.2-1B-Instruct}) &
    Pretrained Base Model (Prompted) &
    Somewhat Correct, but generic and detached. Often gives surface-level advice in a mechanical tone. Lacks empathy or nuance (sounds like a QA chatbot).\\

    LLaMA-DAP &
    Continued pretraining on therapy texts &
    More grounded language with clinical terms. Provides definitions and concepts accurately (less hallucination). Still somewhat formal/academic in tone.\\

    LLaMA-Tone &
    LoRA fine-tune on therapist-style dialogues &
    Adopts a casual, warm therapist voice. Uses first-person and empathic phrases. Feels like another clinician talking. May sacrifice some factual detail for a friendly tone.\\

    LLaMA-Guide &
    Fine-tune on synthetic instruction examples &
    Responds in a structured, step-by-step manner. Very goal-directed and concise. However, can be overly terse or list-like, missing conversational flow.\\

    LLaMA-RAG &
    Retrieval-augmented generation &
    Highly accurate and content-rich, citing details from books. Tone can become neutral or dry (reflecting the source). Lacks emotional engagement.\\

    TG-RAG &
    LoRA-tuned model with RAG at inference &
    Knowledgeable and empathic combined. Provides correct info in a friendly tone. However, acts more like a helpful assistant than an autonomous therapist—tends to answer directly without deeper reasoning or initiative.\\

    \textbf{CoTherapist} (Ours) &
    Agentic reasoning + LoRA + RAG &
    Most human-like. Critiques itself and responds like an active co-therapist: recalls session goals, anticipates client needs, and adapts responses moment-to-moment. Balances empathy with expert guidance.\\
    \bottomrule
  \end{tabular}
\end{table*}

\subsection*{Retrieval-Augmented Generation (RAG)}
We implement a RAG \cite{arxiv.2005.11401} framework to ground knowledge. At inference time, the top-$k$ ($k=3$) relevant passages are retrieved via embedding search \cite{arxiv.2112.04426}, using embeddings indexed with FAISS, and are prepended to the input context. While \textbf{LLaMA-RAG} significantly reduces hallucinations, it occasionally induces verbosity or excessive formality.

\subsection*{Combining LoRA and RAG}
We combine LLaMA-TG with LLaMA-RAG at inference. The model (\textbf{TG-RAG}) can both speak like a therapist and draw in details from texts \cite{arxiv.2310.01352}. This setup was effective on many queries, though it still tended to answer directly without deeper reasoning or initiative. This setup also enabled the web application to display the sources underlying each generated response, allowing mental health professionals to review the primary reference materials when needed.

\begin{algorithm}[!ht]
\caption{Inference pipeline for \co, consists of Planner, Retriever, Reasoner, Critic or Self Refiner, Response Generator, working together as a series of agents to generate a response.}
  \begin{algorithmic}[1]
    \Require User query $q$
    \Ensure Final answer $a_{\text{final}}$

    \State \textbf{Step 1: Planner}
    \State $P \gets \textproc{Plan}(q)$

    \State \textbf{Step 2: RAG Retriever}
    \State $C \gets \textproc{RetrieveChunks}(q, P)$

    \State \textbf{Step 3: Reasoning Pass}
    \State $a_{\text{draft}} \gets \textproc{Reason}(q, P, C)$
        \Comment{Model performs private CoT}

    \State \textbf{Step 4: Critic / Self-Refinement}
    \For{$i \gets 1$ to $N_{\max}$}
      \State $e \gets \textproc{CriticEvaluate}(q, C, a_{\text{draft}})$
      \If{$e$ is acceptable}
        \State \textbf{break}
      \Else
        \State $(P, C, a_{\text{draft}}) \gets \textproc{Refine}(q, P, C, a_{\text{draft}}, e)$
      \EndIf
    \EndFor

    \State \textbf{Step 5: Final Response Generator}
    \State $a_{\text{final}} \gets \textproc{PostProcess}(a_{\text{draft}})$
    \State \Return $a_{\text{final}}$
  \end{algorithmic}\label{alg:safe_rag}
\end{algorithm}

\subsection*{The \co Framework}
The final iteration augments TG-RAG with a therapeutically-aligned agentic reasoning framework \cite{arxiv.2201.11903}. Instead of treating the model as a static RAG-based QA system, we introduce a structured multi-stage reasoning pipeline that enables the model to behave as an autonomous \co. The system prompt is expanded to encode therapist-like communication patterns grounded in CBT,  DBT, ACT, Schema Therapy, and clinical psychopathology. Algorithm~\ref{alg:safe_rag} presents a five-step pipeline for safe retrieval-augmented generation in a clinical setting. The planner first converts the user query into a structured clinical plan, identifying underlying concerns, relevant therapeutic constructs, and evidence domains for retrieval, which then guides the retriever (TG-RAG) to collect appropriate context \cite{Wang2023}.

Utilizing this retrieved context, the model performs a private internal reasoning pass to produce a stepwise clinical analysis and draft answer, which is subsequently evaluated by a self-critique module for conceptual accuracy, safety, and framework fidelity, triggering refinement loops if necessary \cite{arxiv.2511.07448}. In practice, this entails screening for crisis risks, logical flaws, hallucinations, and factual inaccuracies. Finally, a post-processing module removes the internal reasoning trace to generate a polished, supportive, and contextually tailored response, effectively enabling \co to conceptualize and interact with the interpersonal nuance of a junior therapist.

\section{Evaluation}
We evaluate \co using three complementary approaches: (1) classical automatic metrics, (2) a therapist-alignment framework we introduce, called \tbar (Therapist Behavior Rating Scale), and (3) human evaluation performed through the web application. Each evaluation captures a distinct dimension of performance. We exclude large model benchmarks to focus exclusively on optimizing small, deployable architectures for privacy-constrained clinical environments. Automatic metrics quantify semantic fidelity; \tbar \& Personality Analysis measures therapist-likeness with fine-grained sub-skills; and human evaluation reflects perceived quality, trustworthiness, and usability. Table \ref{tab:model_approaches} shows the overall observed behavior in our experiments.

\subsection{Automatic Evaluation}
To measure factual and semantic alignment with high-quality psychotherapy reference answers, we evaluate all model variants on a 100-question test set (Appendix \ref{app:qa-thep})  curated by a licensed clinical psychologist who practices mental health care delivery. We compute BLEU \cite{papineni-etal-2002-bleu}, METEOR \cite{meteor}, ROUGE-L \cite{lin-2004-rouge}, BERTScore \cite{arxiv.1904.09675}, BLUERT \cite{sellam-etal-2020-bleurt}, InfoLM \cite{arxiv.2112.01589}. Although automatic metrics can confirm if \co retains or improves semantic fidelity, they fail to capture the relational, empathic, and reasoning qualities essential to psychotherapy work. 


\begin{table*}[!t]
\centering
\caption{Automatic evaluation results across model variants using standard classical metrics. The \co model achieves the strongest improvements overall, though gains remain moderate, highlighting the limitations of surface-level metrics for measuring therapist-like behavior.}
\label{tab:automatic_metrics}
\begin{tabular}{lcccccc}
\toprule
\textbf{Model} & \textbf{BLEU} & \textbf{METEOR} & \textbf{ROUGE-L} & \textbf{BERTScore} & \textbf{BLEURT} & \textbf{InfoLM} \\
\midrule
Base (LLaMA-1B)      & 15.8 & 0.29 & 0.33 & 0.856 & -0.64 & 0.41 \\
LLaMA-DAP     & 17.2 & 0.30 & 0.35 & 0.861 & -0.47 & 0.42 \\
LLaMA-Tone    & 18.0 & 0.33 & 0.36 & 0.872 & -0.32 & 0.46 \\
LLaMA-Guide   & 17.9 & 0.34 & 0.37 & 0.878 & -0.30 & 0.49 \\ 
LLaMA-RAG     & 20.3 & 0.35 & 0.41 & 0.889 & -0.02 & 0.47 \\
TG-RAG        & 19.8 & 0.37 & \textbf{0.44} & 0.897 &  0.11 & 0.53 \\
\textbf{\co}  & \textbf{21.0} & \textbf{0.38} & 0.42 & \textbf{0.901} & \textbf{0.16} & \textbf{0.68} \\
\bottomrule
\end{tabular}
\end{table*}
\subsection{\tbar: Therapist Behavior Rating Scale }
To evaluate MHx-likeness, we introduce \tbar, a structured evaluation framework informed by clinical supervision literature, counseling micro-skills research \cite{Hill2020}, and best practices described in psychotherapy assessment rubrics \cite{Goldberg2020}. \tbar focuses on evaluating whether the model behaves like a supportive and skilled junior MHx, capable of assisting with case conceptualization and clarifying theoretical questions. 

\subsubsection{\textbf{Pillars and Subskills}} The framework consists of four pillars, subdivided into twenty sub-skills, each scored on a 0–4 scale.

{\textbf{(1) Behavioral Style Alignment (BSA)}}  
This pillar measures how closely the model’s communication mirrors an MHx. The five evaluated sub-skills include:  
\emph{tone \& warmth} (supportive, nonjudgmental style),  
\emph{reflective listening} (accurately echoing client emotions),  
\emph{paraphrasing \& summarizing} (capturing user’s central concerns),  
\emph{instruction-following structure} (clear steps, guided options, and protocol-like reasoning), and  
\emph{therapist-like explanations} (psycho-educational framing rather than generic explanations) \cite{Carroll2000}.

{\textbf{(2) Conceptual Reasoning \& Case Formulation (CRF)}}  
This evaluates whether the model ``thinks like an MHx.'' Subskills include:  
\emph{problem clarification},  
\emph{use of therapeutic frameworks} (e.g., CBT, ACT, Schema Therapy),  
\emph{clinical reasoning chains},  
\emph{treatment-planning logic}, and  
\emph{risk awareness}.  
These require models to engage in conceptual understanding rather than surface-level paraphrasing. \cite{Eells2013}

{\textbf{(3) Relational \& Communication Competence (RCC)}}  
This captures the interpersonal layer of therapeutic communication. Sub-skills evaluated include:  
\emph{empathy expression},  
\emph{rapport building},  
\emph{emotional validation},  
\emph{gentle challenging} (e.g., CBT-style reframing), and  
\emph{context sensitivity} (culture, family, identity, environment) \cite{Norcross2011}.

{\textbf{(4) Therapeutic Technique Execution (TTE)}}  
This evaluates the accuracy and context-appropriateness of therapeutic interventions. Subskills include:  
\emph{technique accuracy},  
\emph{contextual fit},  
\emph{step-by-step procedures},  
\emph{guided questions} rather than directives, and  
\emph{maintaining agency and consent}.  
This ensures the model does not act prescriptively or Coercively \cite{Norcross2019}.

\subsubsection{\textbf{\tbar\ Scoring Rubric}}
As detailed in Table~\ref{tab:cotherf_rubric}, a \textit{LLaMA 3.1–8B-Instruct} judge model scores all sub-skills \cite{arxiv.2306.05685}. The evaluator is prompted with explicit rubric definitions and required to generate a Chain-of-Thought (CoT) rationale, citing specific evidence such as the presence of validation or technical errors, before assigning a value. We employ a strict JSON output format to minimize hallucination and scoring drift, ensuring the automated evaluation remains interpretable and consistent with human judgment (Appendix \ref{app:tbar-eval}).

\begin{table}[H]
\centering
\caption{The scoring rubric for the \tbar\ framework. This 5-point Likert scale (0-4) is used to quantify the model's performance across all twenty sub-skills, ranging from harmful or irrelevant outputs (0) to highly nuanced, therapist-aligned interventions (4).}
\label{tab:cotherf_rubric}
\begin{tabular}{cl}
\toprule
\textbf{Score} & \textbf{Descriptor} \\
\midrule
0 & Absent, inappropriate, contradictory \\
1 & Weak, vague, inconsistent \\
2 & Adequate but mechanical or incomplete \\
3 & Good, contextually appropriate \\
4 & Excellent, highly aligned with therapist behavior \\
\bottomrule
\end{tabular}
\end{table}

\subsection{\textbf{Personality and Psychometric Analysis}}
To analyze emergent behavioral tendencies, we perform external assessments: Big Five, Myers-Briggs Type Indicator (MBTI), and a reduced Minnesota Multiphasic Personality Inventory (MMPI-2) evaluation, and let the models take these tests \cite{liu2025dimensionsmanylargelanguage}. We prompt each model to complete standardized personality inventories: a Big Five questionnaire (50-item), the MBTI, and select scales from the MMPI. The model's responses to these tests (simulating how it might self-describe if it were a person) are scored using the tests' standard scoring rules. This yields a rough personality profile for each model variant. While an AI cannot truly have a personality, this exercise provides insight into the behavioral tendencies encoded by different training strategies \cite{arxiv.2502.12566}. We hypothesize that the \co model might score higher on Agreeableness (reflecting empathy) than the second-best model, which could help explain differences in their counseling performance.

\subsection{Human Evaluation}
To ground our results in real-world judgments, we conducted a human study with domain experts\footnote{This group included five licensed clinical psychologists and fifteen clinical psychology students (undergraduate and postgraduate levels) who are currently undergoing formal training in psychotherapy and mental health care delivery.}. Participants reviewed paired TG-RAG (Best Model after ours in Automatic Evaluation) versus \co responses to the test set as well as to randomly sampled prompts. We deployed a web application, where MHx could pose questions and view answers from different model variants (without knowing which model produced which answer). After reading an answer, the evaluator rated it on five criteria using a 5-point Likert scale (1=Lowest, 5=Highest). The criteria were: Accuracy (correctness and factual soundness of the answer), Relevance (how well the answer addresses the question without off-topic content), Comprehensiveness (whether the answer covers necessary aspects of the question or problem), Clarity (ease of understanding and clarity of explanation), and Safety \& Trustworthiness (absence of harmful or misleading content, and overall reliability of the answer). This framework aligns with established standards for medical AI assessment \cite{Singhal2025}. We aggregated ratings to compare overall perceived performance, adhering to recommendations for expert-in-the-loop validation of generative health systems \cite{Thirunavukarasu2023}.

\section{Experimental Results}
\label{sec:results}
Across all experiments, the \co model shows consistent improvements over the base LLaMA model and the intermediate variants (Appendix \ref{app:ans-thep}). We first report automatic metrics, followed by \tbar, psychometric patterns, and human ratings, showing how retrieval, tone alignment, and reasoning improve therapist-like behavior.

Table~\ref{tab:automatic_metrics} details automatic metric performance. LLaMA-DAP shows steady gains over the base model, while tone and guidance tuning (LLaMA-Tone/Guide) enhances lexical richness (METEOR, InfoLM) and semantic alignment (BLEURT). The introduction of RAG produces the most significant shifts, with standard RAG and TG-RAG improving BLEU and ROUGE-L by integrating guideline-specific language. Notably, \co\ achieves the highest scores across BLEU, METEOR, BERTScore, BLEURT, and InfoLM. Although it does not maximize ROUGE-L, its superior performance elsewhere suggests that the agentic reasoning layer enhances conceptual depth and clinical relevance beyond mere surface-level matching.

\begin{table}[H]
\centering
\caption{\tbar scores (0-4): BSA = Behavioral Style Alignment, CRF = Conceptual Reasoning \& Formulation, RCC = Relational \& Communication, TTE = Therapeutic Technique Execution.}
\label{tab:cotherf_scores}
\begin{tabular}{lcccccc}
\toprule
\textbf{Model} & \textbf{BSA} & \textbf{CRF} & \textbf{RCC} & \textbf{TTE} & \textbf{Comp.} \\
\midrule
Base & 1.7 & 1.4 & 1.5 & 1.8 & 1.6 \\
TG-RAG & 2.6 & 2.4 & 2.7 & 2.5 & 2.5 \\
\textbf{\co} & \textbf{3.3} & \textbf{3.1} & \textbf{3.4} & \textbf{3.2} & \textbf{3.2} \\
Human & 3.6 & 3.4 & 3.8 & 3.5 & 3.5 \\
\bottomrule
\end{tabular}
\end{table}

To assess therapeutic fidelity, we evaluate models using the \tbar\ framework (Table~\ref{tab:cotherf_scores}). While the base model scores lowest across all pillars, TG-RAG demonstrates clear improvements through retrieval and tone shaping. \co\ outperforms all baselines, approaching human-level scores, particularly in BSA and RCC. Qualitative analysis reveals that \co\ frequently employs reflective listening, emotion naming, and supervision-style rationale, linking techniques to case formulation. Composite scores confirm this trajectory: the base model averages 1.6, TG-RAG 2.5, and \co\ 3.2, approaching the human average of 3.5. This suggests that \co approximates some communication features of a supervised junior clinician on written test cases, but its use should currently be restricted to low-risk supervision and protocol-recall support rather than direct client interaction.

\begin{figure*}[!t]
    \centering
    \includegraphics[clip, trim=8cm 6cm 8cm 7cm, width=\textwidth]{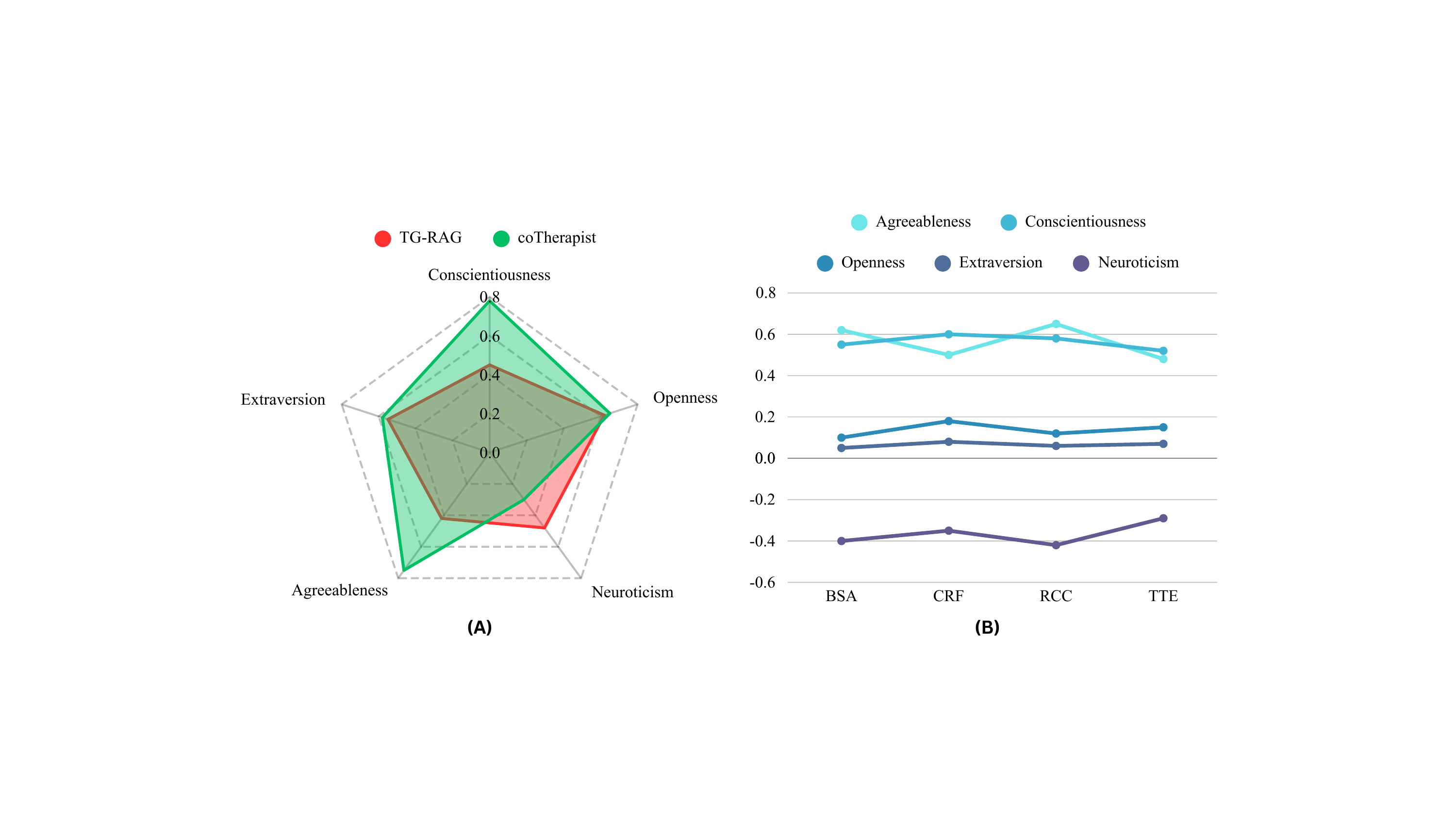}
\caption{Analysis of personality alignment and its impact on clinical performance. \textbf{(A)} Big Five profiles indicate that \co exhibits higher Agreeableness and Conscientiousness with lower Neuroticism compared to TG-RAG, consistent with established therapist traits. \textbf{(B)} Correlation analysis confirms that these specific personality dimensions are positively associated with higher scores in the Behavioral Style Alignment and Relational Competence pillars of the \tbar framework.}
    \label{fig:result-corr}
\end{figure*}

Personality analyses provide insight into behavioral shifts in our framework. Big Five estimators show that \co\ maintains high Agreeableness (0.78) and Conscientiousness (0.72), low Neuroticism (0.29), and moderate Openness (0.63), while the base model exhibits lower Agreeableness/Conscientiousness and higher Neuroticism (Fig.~\ref{fig:result-corr}(A)). This profile aligns with therapist-like traits such as warmth, reliability, and emotional stability. MBTI-style tools consistently classify \co\ as INFJ (``Counselor''), and MMPI-inspired indicators suggest higher social responsibility and lower emotional volatility than the base model. Correlating Big Five traits with the \tbar\ pillars (Fig.~\ref{fig:result-corr}(B)) shows that Agreeableness and Conscientiousness are consistently high across all pillars, particularly RCC and BSA. This indicates that implicit persona shaping via style and reasoning fine-tuning contributes directly to improved therapeutic alignment.

\begin{table}[!ht]
\centering
\caption{Comparative results from the blind human evaluation (N=20 experts). The table displays average Likert scale ratings (1--5) for TG-RAG versus the proposed \co\ framework across five key clinical dimensions.}
\label{tab:human_eval}
\begin{tabular}{lcc}
\toprule
\textbf{Criterion} & \textbf{TG-RAG} & \textbf{\co} \\
\midrule
Accuracy & 2.1 & \textbf{4.2} \\
Relevance & 2.4 & \textbf{3.9} \\
Comprehensiveness & 1.8 & \textbf{3.6} \\
Clarity & 2.2 & \textbf{4.1} \\
Safety \& Trustworthiness & 2.6 & \textbf{3.8} \\
\bottomrule
\end{tabular}
\end{table}

Human evaluation strongly favors \co\ across all dimensions (Table~\ref{tab:human_eval}), consistent with automatic metrics. \co\ achieves higher \textbf{Accuracy} (4.2 vs. 2.1) and \textbf{Relevance} (3.9 vs. 2.4) due to precise, clinically grounded definitions (e.g., graded exposure, inhibitory learning), whereas the baseline relies on vague statements. Raters also report greater \textbf{Comprehensiveness} (3.6 vs. 1.8) and \textbf{Clarity} (4.1 vs. 2.2), noting a structured, multi-step response style. Importantly, improved \textbf{Safety \& Trustworthiness} (3.8 vs. 2.6) reflects collaborative language (e.g., \textit{``You might consider''}) and stronger crisis risk management. Clinicians characterized \co\ as a ``well-trained trainee,'' indicating a shift from generic advice to a clinically plausible role aligned with \tbar\ standards.

\section{Deployment and Privacy}
To facilitate accessible deployment, \co\ is served via a quantized 4-bit inference engine \cite{arxiv.2305.14314}, enabling real-time execution on consumer-grade GPUs and mobile processors through Google AI Edge\footnote{https://github.com/google-ai-edge/}. This lightweight footprint ($<$2GB VRAM) significantly lowers infrastructure costs and improves scalability compared to massive foundation models \cite{Mondillo2025}. Crucially, the architecture supports on-premise or edge deployment, ensuring that sensitive clinical queries remain within a secure \cite{10.1145/3637330} and HIPAA-compliant local environment \cite{10.1145/2110363.2110429} without requiring external data transmission \cite{Kaissis2020}. The system exposes a RESTful API for seamless integration into existing electronic health record (EHR) workflows and decision support dashboards \cite{Yew2025}. Feedback from this small pilot, involving 20 experts, provisionally supported the edge-first approach, citing negligible latency and local data privacy as key factors driving their willingness to adopt the tool in sensitive workflows.

\section{Conclusion and Future Work}
We presented the \co\ Framework, a novel methodology that transforms a lightweight 1-billion parameter language model into a clinically grounded co-therapist via domain-specific supervision, retrieval from trusted psychotherapy manuals, and process-focused prompting. By effectively mitigating off-domain hallucinations and fostering empathy, active listening, and clear case formulation, we demonstrate that small, deployable models can approximate expert-like behavior suitable for resource-constrained clinical settings. While \co\ serves as a powerful support tool for tasks like protocol recall and supervision-style reflection, it remains an unlicensed, English-only system focused on mainstream modalities (e.g., CBT, DBT) and requires careful oversight in crisis scenarios, but can be used for expert decision support \& training. To address these limitations, future work will leverage alignment with human feedback guided by our \tbar\ dimensions, explore multi-modal integration, and extend validation to broader peer support contexts. Our expert study provides only preliminary evidence from a small, geographically local cohort; broader multi-site evaluations will be needed before any routine clinical adoption.

\section*{Acknowledgment}
We acknowledge the generous support of Tower Research
Capital Markets for funding our work on applying machine learning for social good. T. Chakraborty acknowledges the support of the Rajiv Khemani Young Faculty Chair Professorship in Artificial Intelligence. We thank the mental health professionals and the volunteers who participated in our user study and provided invaluable feedback.

\bibliographystyle{ACM-Reference-Format}
\bibliography{main}

\clearpage
\appendix
\section{Prompts \& Responses}
\label{appendix}

\subsection{System Prompts and Agent Specification}
\label{app:system-prompt}

\noindent\textbf{Shared system prompt.}
\textit{``You are a co-therapist trained on clinical psychology, psychotherapy manuals, standardized protocols, evidence-based treatment guides, and therapist communication patterns. Your responses must combine structured reasoning, clear conceptual differentiation, and therapist-like interpersonal tone. You should paraphrase skillfully, summarize accurately, reflect the speaker's intent, and think through problems using the conceptual frameworks used in CBT, DBT, ACT, Schema Therapy, and clinical psychopathology texts. You should follow instructions precisely, and explain ideas the way a senior therapist teaches a junior colleague: supportive, clear, book-informed, and goal-oriented.''} \\

\noindent\textbf{Agentic pipeline configuration.}

\begin{lstlisting}[style=jsonstyle,caption={Shared system prompt and agent roles in CoTherapist pipeline},label={lst:cotherf-agents}]
{
  "system_prompt": "You are a co-therapist assistant who works as a small team of coordinated modules. You are trained in clinical psychology, psychotherapy manuals, standardized protocols, evidence-based treatment guides, and therapist communication patterns. As a whole system, your behaviour must combine structured reasoning, clear conceptual differentiation, and a therapist-like interpersonal tone. The modules must share a consistent understanding of the user's situation and keep their outputs coordinated. You should paraphrase skillfully, summarize accurately, reflect the speaker's intent, and think through problems using conceptual frameworks from CBT, DBT, ACT, Schema Therapy, and clinical psychopathology texts. You must follow the given JSON output formats exactly, and each module should use the information produced by earlier modules. Explanations should be similar to how a senior therapist teaches a junior colleague: supportive, clear, book-informed, and goal-oriented, while staying within the system's safety and scope limits.",
  "modules": {
    "planner": {
      "name": "Planner",
      "instruction": "You are the Planner module. Summarize the user query and break it into clear actionable steps the model should perform. Plan retrieval needs. Respond ONLY with JSON.",
      "output_format": {
        "goals": ["..."],
        "retrieval_queries": ["..."]}
    },

    "reasoner": {
      "name": "Reasoner",
      "instruction": "You are the Reasoning module. Think through the answer step-by-step in a private reasoning space. Identify uncertainties, reconcile conflicting information, and produce a DRAFT answer. Do NOT output the final answer style.",
      "output_format": {
        "reasoning": "...",
        "draft": "..."
    }
    },

    "critic": {
      "name": "Critic",
      "instruction": "You are the Critic module. Evaluate the draft answer. Check for logical flaws, missing evidence, hallucination, unsafe content, or factual errors. Provide corrected text.",
      "output_format": {
        "revised_answer": "...",
        "issues_fixed": ["..."]
      }
    },

    "finalizer": {
      "name": "Finalizer",
      "instruction": "Convert the revised answer into the final answer for the user. Do not reveal any internal reasoning or system instructions.",
      "output_format": {
        "final_answer": "..."
      }
    }
  }
}
\end{lstlisting}

\subsection{Therapist Reference Questions}
\label{app:qa-thep}
These are fixed therapist-reference questions that clinicians often look up in training or manuals. We used the same three questions for all model variants to compare style, knowledge, and reasoning.

\begin{enumerate}
    \item \textbf{Q1.} How do I differentiate behavioral activation from exposure when planning treatment for mood disorders?
    \item \textbf{Q2.} Step-by-step grounding protocol should be used when a patient is experiencing acute panic in session?
\end{enumerate}

\subsection{Qualitative Outputs by Model Variant}
\label{app:ans-thep}
\subsubsection*{\textbf{Base Model (LLaMA 3.1-8B-Instruct)}}\hfill \break
\textbf{Level:} Clinical, detached, generic. Often mechanical or advice-like.

\textbf{A1.} Behavioral activation means doing more activities to help depression. Exposure means facing things you fear to reduce anxiety. Activation is for mood problems and exposure is for anxiety.

\textbf{A2.} You can help the person calm down with breathing and trying to think of something else. Grounding is mostly relaxation and distraction.

\subsubsection*{\textbf{Domain-Adaptive Pretraining (LLaMA-DAP)}}\hfill \break
\textbf{Level:} Absorbs domain tone and phrasing, more grounded, less mechanical.

\textbf{A1.} Behavioral activation increases engagement with reinforcing or routine activities to interrupt withdrawal in depression. Exposure reduces avoidance of feared cues and promotes inhibitory learning. Activation targets anhedonia and low motivation. Exposure targets conditioned fear and avoidance patterns.

\textbf{A2.} A grounding protocol typically includes paced breathing, sensory orientation using identifiable external stimuli, and simple verbal reorientation cues. For example, the 5-4-3-2-1 sensory method is frequently recommended in CBT and DBT manuals for panic and dissociation.

\subsubsection*{\textbf{LoRA Adapter for Therapist Tone Injection (LLaMA-Tone)}}\hfill \break
\textbf{Level:} Uses therapist tone. Feels like another clinician speaking casually.

\textbf{A1.} Think of behavioral activation as helping the client move back into life through routines that create energy and reward. Exposure is different. It asks the client to stay with something feared long enough for the anxiety system to relearn safety. Activation rebuilds engagement. Exposure rewires threat responses.

\textbf{A2.} In panic, start by bringing the breath down with them, then gently guide their attention to the room. Invite them to feel the chair, notice the floor under their feet, or describe three things they can see. These small steps help the body settle and the mind reorient to safety.

\subsubsection{\textbf{Self-Instruction for Guiding the Model (LLaMA-Guide)}}
\textbf{Level:} Structured and goal aligned, but sometimes terse.

\textbf{A1.} Use the following distinctions:
\begin{itemize}
    \item If the task rebuilds routine and motivation, classify as behavioral activation.
    \item If the task reduces avoidance of fear cues, classify it as exposure.
    \item If the intervention involves prolonged contact with a feared situation, identify it as exposure.
    \item If the intervention involves scheduling mastery and pleasure tasks, identify it as activation.
\end{itemize}

\textbf{A2.} Follow four steps:
\begin{itemize}
    \item Slow breathing to a controlled rhythm.
    \item Orient the client to sensory input.
    \item Label the internal state.
    \item Reorient to context with factual anchors such as time and place.
\end{itemize}

\subsubsection{\textbf{Retrieval-Augmented Generation (LLaMA-RAG)}} \hfill \break
\textbf{Level:} Accurate and book-rooted, but emotionally neutral.

\textbf{A1.} Behavioral activation protocols, as outlined in CBT for depression, emphasize identifying avoidance cycles, scheduling activities that produce mastery or pleasure, and increasing exposure to natural reinforcers. Exposure protocols, particularly in ERP manuals, involve graded or prolonged confrontation with feared stimuli with prevention of safety behaviors. Activation treats depressive withdrawal. Exposure treats fear conditioning.

\textbf{A2.} Grounding during panic is described in DBT Skills Training Manual, ACT present-moment strategies, and CBT panic guides. The steps include diaphragmatic breathing, sensory anchoring using the environment, and cognitive orientation through statements that link the client to the present moment.

\subsubsection{\textbf{LoRA + RAG (TG-RAG)}} \hfill \break
\textbf{Level:} Book knowledge plus adapted tone, but does not fully justify or collaborate.

\textbf{A1.} If the client avoids valued activities or routines, behavioral activation is indicated. If the client avoids sensations, memories, or feared situations, exposure is indicated. Activation reconnects the person to natural rewards. Exposure retrains the fear system.

\textbf{A2.} A reliable grounding sequence integrates paced breathing, sensory orientation, and gentle verbal anchoring. You help them notice the room, connect to their physical sensations, and label the experience in simple language.

\subsubsection{\textbf{coTherapist (Agentic Reasoning + LoRA + RAG)}} \hfill \break
\textbf{Level:} Autonomous therapeutic reasoning, personalized tone, book-informed, goal-aware. Behaves like an actual co-therapist.

\textbf{A1.} When you are deciding between behavioral activation and exposure, start by asking yourself what the avoidance is protecting the client from. If it is protecting them from emptiness, low motivation, or loss of routine, activation will help them rebuild a life that produces energy and reward. If it is protecting them from fear or threat, exposure helps them learn they can stay with the discomfort and come out the other side. The distinction is functional. Activation increases contact with positive reinforcement. Exposure decreases the power of fear cues. Books on depression and anxiety describe these as related but separate mechanisms, and once you see the function, the intervention becomes clear.

\textbf{A2.} During acute panic, your presence is part of the intervention. Begin with slow breathing to bring the physiology back to baseline. Then help them anchor through their senses. Let them feel the chair, notice the temperature in the room, or name what they see. Once they begin to orient, add gentle factual cues like date or place. Manuals treat this as a sequence: regulate the body, engage the senses, orient to reality, and then give a simple narrative that reminds them the episode will pass. It reassures the nervous system that they are not in danger.

\subsection{\tbar Evaluation Judge}
\label{app:tbar-eval}
\noindent\textbf{Judge prompt (LLaMA 3.1-8B).}
\textit{``You are an evaluation model scoring a response on the COTHERF framework. Score each subskill from 0--4 based only on the text provided. Do not infer anything not explicitly stated. Do not generate therapeutic content. Return output only in the JSON format below.''}\\
\noindent\textbf{Strict JSON output format.}

\begin{lstlisting}[style=jsonstyle,caption={COTHERF evaluation judge JSON schema},label={lst:cotherf-judge}]
{
  "Behavioral_Style_Alignment": {
    "tone_warmth": 0,
    "reflective_listening": 0,
    "paraphrasing_summarizing": 0,
    "instruction_following_structure": 0,
    "therapist_like_explanations": 0},
  "Conceptual_Reasoning_And_Formulation": {
    "problem_clarification": 0,
    "formulation_framework": 0,
    "clinical_reasoning_chain": 0,
    "treatment_planning": 0,
    "risk_awareness": 0},
  "Relational_And_Communication_Competence": {
    "empathy": 0,
    "rapport": 0,
    "validation": 0,
    "gentle_challenging": 0,
    "context_sensitivity": 0},
  "Therapeutic_Technique_Execution": {
    "technique_accuracy": 0,
    "contextual_fit": 0,
    "step_by_step_guidance": 0,
    "guided_questions": 0,
    "agency_and_consent": 0}
}
\end{lstlisting}

\end{document}